\documentclass[twocolumn,conference, 10pt, a4paper]{IEEEtran}
\usepackage[latin9]{inputenc}
\usepackage{amssymb}
\usepackage{graphicx}
\usepackage[unicode=true,
 bookmarks=true,bookmarksnumbered=true,bookmarksopen=true,bookmarksopenlevel=1,
 breaklinks=false,pdfborder={0 0 0},pdfborderstyle={},backref=false,colorlinks=false]
 {hyperref}
\hypersetup{pdftitle={Your Title},
 pdfauthor={Your Name},
 pdfpagelayout=OneColumn, pdfnewwindow=true, pdfstartview=XYZ, plainpages=false}

\makeatletter

\providecommand{\tabularnewline}{\\}

\usepackage[caption=false,font=footnotesize]{subfig}
\usepackage{microtype}
\usepackage{flushend}

\usepackage{amsbsy}
\usepackage{algorithmic}
\usepackage[english]{babel}
\usepackage[T1]{fontenc}

\makeatother

\begin{document}

\title{Faster Training of Very Deep Networks Via $p$-Norm Gates}

\author{\IEEEauthorblockN{Trang~Pham, Truyen~Tran, Dinh~Phung, Svetha~Venkatesh}\IEEEauthorblockA{Center for Pattern Recognition and Data Analytics\\
Deakin University, Geelong Australia\\
Email: \{phtra, truyen.tran, dinh.phung, svetha.venkatesh\}@deakin.edu.au}}

\maketitle
\begin{abstract}
A major contributing factor to the recent advances in deep neural
networks is structural units that let sensory information and gradients
to propagate easily. Gating is one such structure that acts as a flow
control. Gates are employed in many recent state-of-the-art recurrent
models such as LSTM and GRU, and feedforward models such as Residual
Nets and Highway Networks. This enables learning in very deep networks
with hundred layers and helps achieve record-breaking results in vision
(e.g., ImageNet with Residual Nets) and NLP (e.g., machine translation
with GRU). However, there is limited work in analysing the role of
gating in the learning process. In this paper, we propose a flexible
$p$-norm gating scheme, which allows user-controllable flow and as
a consequence, improve the learning speed. This scheme subsumes other
existing gating schemes, including those in GRU, Highway Networks
and Residual Nets as special cases. Experiments on large sequence
and vector datasets demonstrate that the proposed gating scheme helps
improve the learning speed significantly without extra overhead.
\end{abstract}

\IEEEpeerreviewmaketitle{}

\global\long\def\xb{\boldsymbol{x}}
\global\long\def\yb{\boldsymbol{y}}
\global\long\def\eb{\boldsymbol{e}}
\global\long\def\zb{\boldsymbol{z}}
\global\long\def\hb{\boldsymbol{h}}
\global\long\def\ab{\boldsymbol{a}}
\global\long\def\bb{\boldsymbol{b}}
\global\long\def\cb{\boldsymbol{c}}
\global\long\def\sigmab{\boldsymbol{\sigma}}
\global\long\def\gammab{\boldsymbol{\gamma}}
\global\long\def\alphab{\boldsymbol{\alpha}}
\global\long\def\rb{\boldsymbol{r}}
\global\long\def\fb{\boldsymbol{f}}
\global\long\def\ib{\boldsymbol{i}}
\global\long\def\thetab{\boldsymbol{\theta}}

\section{Introduction}

Deep neural networks are becoming the method of choice in vision \cite{he2015delving},
speech recognition \cite{hinton2012deep} and NLP \cite{sutskever2014sequence,kumar2015ask}.
Deep nets represent complex data more efficiently than shallow ones
\cite{bengio2009learning}. With more non-linear hidden layers, deep
networks can theoretically model functions with higher complexity
and nonlinearity \cite{choromanska2015loss}. However, learning standard
feedforward networks with many hidden layers is notoriously difficult
\cite{larochelle2009exploring}. Likewise, standard recurrent networks
suffer from vanishing gradients for long sequences \cite{hochreiter2001gradient}
making gradient-based learning ineffective. A major reason is that
many layers of non-linear transformation prevent the data signals
and gradients from flowing easily through the network. In the forward
direction from data to outcome, a change in data signals may not lead
to any change in outcome, leading to the poor credit assignment problem.
In the backward direction, a large error gradient at the outcome may
not be propagated back to the data signals. As a result, learning
stops prematurely without returning an informative mapping from data
to outcome.

There have been several effective methods to tackle the problem. The
first line of work is to use non-saturated non-linear transforms such
as rectified linear units (ReLUs) \cite{glorot2011deep,he2015delving,goodfellow2013maxout},
whose gradients are non-zero for a large portion of the input space.
Another approach that also increases the level of linearity of the
information propagation is through \emph{gating} \cite{hochreiter1997long,graves2013generating}.
The gates are extra control neural units that let part of information
pass through a channel. They are learnable and have played an important
role in state-of-the-art feedforward architectures such as Highway
Networks \cite{srivastava2015training} and Residual Networks \cite{he2015deep},
and recurrent architectures such as such as Long Short-Term Memory
(LSTM) \cite{hochreiter1997long,graves2013generating} and Gated Recurrent
Unit (GRU) \cite{cho2014properties}.

Although the details of these architectures differ, they share a common
gating scheme. More specifically, let $\hb_{t}$ be the activation
vector of size $K$ (or memory cells, in the case of LSTM) at computational
step $t$, where $t$ can be the index of the hidden layer in feedforward
networks, or the time step in recurrent networks. The updating of
$\hb_{t}$ follows the following rule:

\begin{equation}
\hb_{t}\leftarrow\alphab_{1}*\tilde{\hb}_{t}+\alphab_{2}*\hb_{t-1}\label{eq:generic}
\end{equation}
where $\tilde{\hb}_{t}$ is the non-linear transformation of $\hb_{t}$
(and the input at $t$ if given), $\alphab_{1},\alphab_{2}\in\left[0,1\right]^{k}$
are gates and $*$ is point-wise multiplication. When $\alphab_{2}>\mathbf{0}$,
a part of the previous activation vector is copied into the new vector.
Thus the update has a nonlinear part (controlled by $\alphab_{1}$)
and a linear part (controlled by $\alphab_{2}$). The nonlinear part
keeps transforming the input to more complex output, whilst the linear
part retains a part of input to pass across layers much easier. The
linear part effectively prevents the gradient from vanishing even
if there are hundreds of layers. For example, Highway Networks can
be trained with more than 1000 layers \cite{srivastava2015training},
which were previous impossible for feedforward networks.

This updating rule opens room to study relationship between the two
gates $\alphab_{1}$ and $\alphab_{2}$, and there has been a limited
work in this direction. Existing work includes Residual Networks with
$\alphab_{1}=\alphab_{2}=\mathbf{1}$, hence $\tilde{\hb}_{t}$ plays
the role of the residual. For the LSTM, there is no explicit relation
between the two gates. The GRU and the work reported in \cite{srivastava2015training}
use $\alphab_{1}+\alphab_{2}=\mathbf{1}$, which leads to less parameters
compared to the LSTM. This paper focuses on the later, and aims to
address the inherent drawback in this linear relationship. In particular,
when $\alphab_{1}$ approaches $\mathbf{1}$ with rate $\lambda$,
$\alphab_{2}$ approaches $\mathbf{0}$ with the same rate, and this
may prevent information from passing too early. To this end we propose
a more flexible $p$-norm gating scheme, where the following relationship
holds: $\left(\alphab_{1}^{p}+\alphab_{2}^{p}\right)^{1/p}=\mathbf{1}$
for $p>0$ and the norm is applied element-wise. This introduces just
one an extra controlling hyperparameter $p$. When $p=1$, the scheme
returns to original gating in Highway Networks and GRUs.

We evaluate this $p$-norm gating scheme on two settings: the traditional
classification of vector data under Highway Networks and sequential
language models under the GRUs. Extensive experiments demonstrate
that with $p>1$, the learning speed is significantly higher than
existing gating schemes with $p=1$. Compared with the original gating,
learning with $p>1$ is 2 to 3 times faster for vector data and more
than 15\% faster for sequential data. 

The paper is organized as follows. Section 2 presents the Highway
Networks, GRUs and the $p$-norm gating mechanism. Experiments and
results with the two models are reported in Section 3. Finally, Section
4 discusses the findings further and concludes the paper.

\section{Methods}

In this section, we propose our $p$-norm gating scheme. To aid the
exposition, we first briefly review the two state-of-the-art models
that use gating mechanisms: Highway Networks \cite{srivastava2015training}
(a feedforward architecture for vector-to-vector mapping) and Gated
Recurrent Units \cite{cho2014properties} (a recurrent architecture
for sequence-to-sequence mapping) in Sections~\ref{subsec:Highway-Networks}
and \ref{subsec:Gated-Recurrent-Units}, respectively. 

\subsubsection*{Notational convention }

\begin{figure}[h]
\begin{centering}
\begin{tabular}{cc}
\includegraphics[bb=350bp 170bp 650bp 460bp,clip,width=0.48\columnwidth]{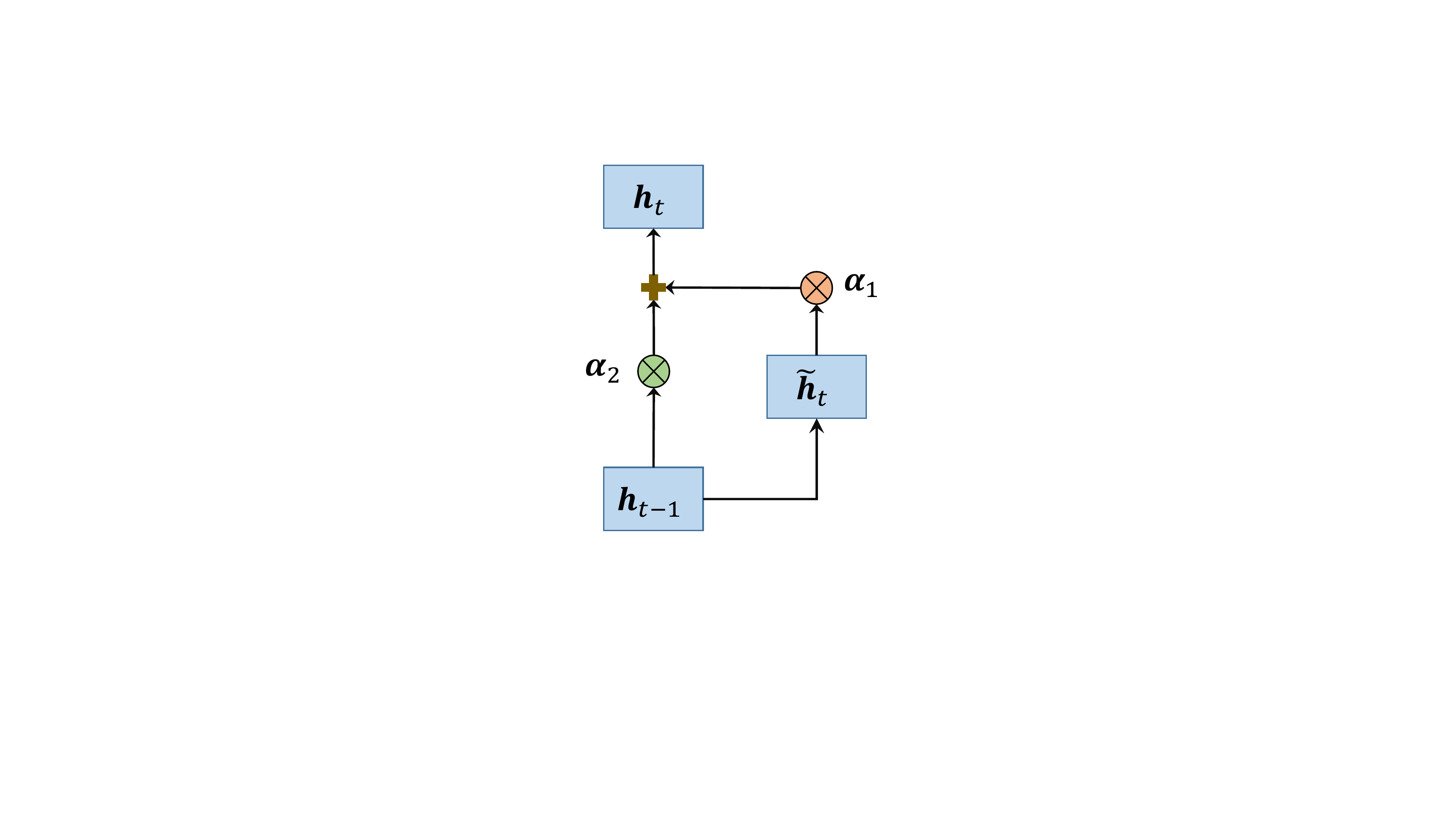} & \includegraphics[bb=350bp 170bp 650bp 460bp,clip,width=0.48\columnwidth]{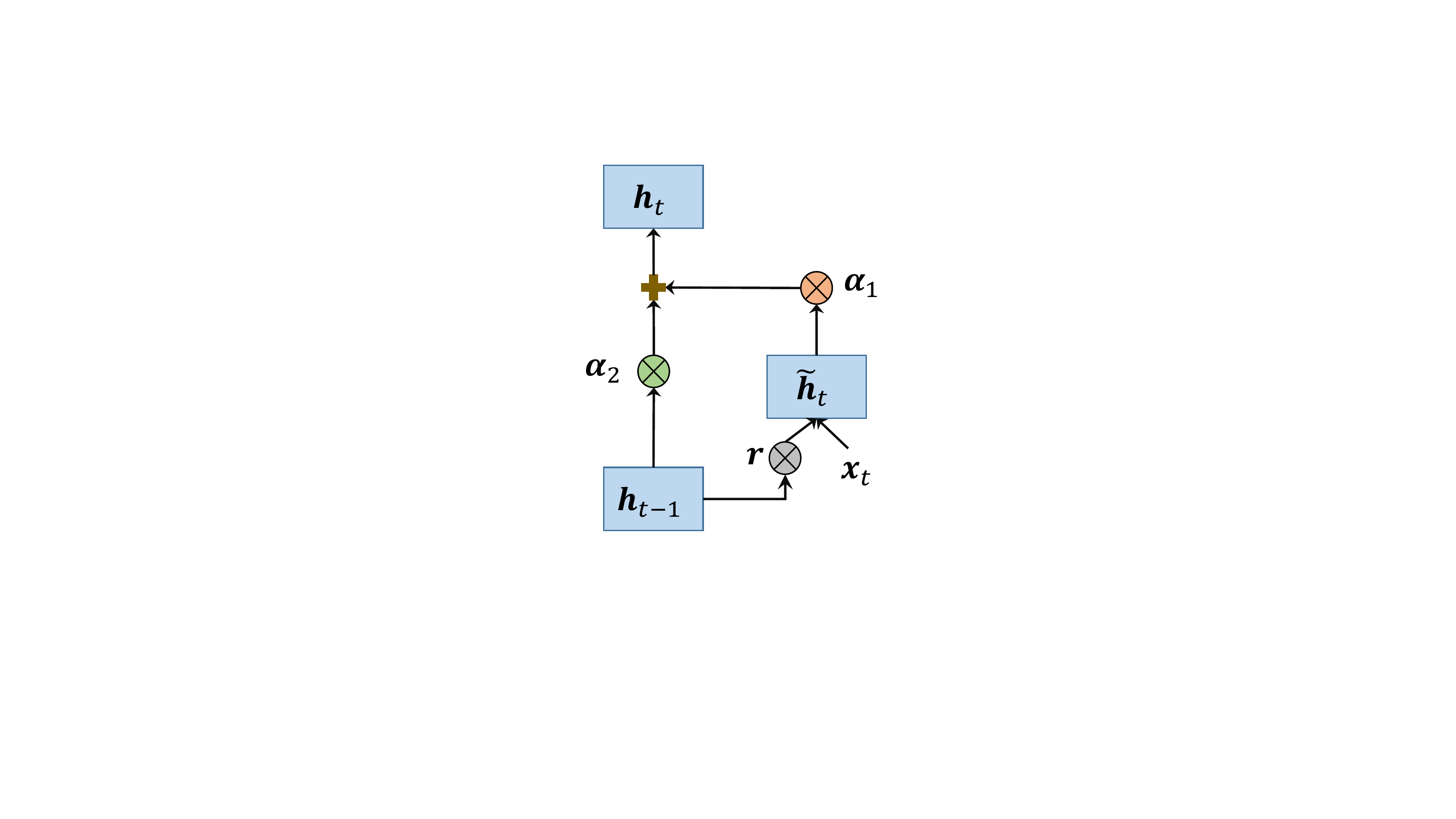}\tabularnewline
(a) Highway Network layer & (b) Gated Recurrent Unit\tabularnewline
\end{tabular}
\par\end{centering}
\caption{Gating mechanisms in Highway Networks and GRUs. The current hidden
state $\protect\hb_{t}$ is the sum of candidate hidden state $\tilde{\protect\hb}_{t}$
moderated by $\protect\alphab_{1}$ and the previous hidden state
$\protect\hb_{t-1}$ moderated by $\protect\alphab_{2}$ \label{fig:highway_GRU}}
\end{figure}

We use bold lowercase letters for vectors and capital letters for
matrices. The sigmoid function of a scalar $x$ is defined as $\sigma(x)=\left[1+\exp(-x)\right]^{-1},\mbox{ }x\in\mathbb{R}$.
With a slight abuse of notation, we use $\sigma\left(\xb\right)$,
where $\xb=\left(x_{1},x_{2},...,x_{n}\right)$ is a vector, to denote
a vector $\left(\sigma\left(x_{1}\right),...,\sigma\left(x_{n}\right)\right)$.
The same rule applies to other function of vector $g(\xb)$. The operator
$*$ is used to denote element-wise multiplication. For both the feedforward
networks and recurrent networks, we use index $t$ to denote the \emph{computational
steps}, and it can be layers in feedforward networks or time step
in recurrent networks. As shown from Fig.~\ref{fig:highway_GRU},
the two architectures are quite similar except for when extra input
$\xb_{t}$ is available at each step. 

\subsection{Highway Networks \label{subsec:Highway-Networks}}

A Highway Network is a feedforward neural network which maps an input
vector $\xb$ to an outcome $y$. A standard feedforward network consists
of $T$ hidden layers where the activation $\hb_{t}\in\mathbb{R}^{k_{t}}$
at the $t^{th}$ layer $\left(t=1,..,T\right)$ is a non-linear function
of its lower layer:

\[
\hb_{t}=g\left(W_{t}\hb_{t-1}+\bb_{t}\right)
\]
where $W_{t}$ and $\bb_{t}$ are the parameter matrix and bias vector
at the $t^{th}$ layer, and $g(\cdot)$ is the element-wise non-linear
transform. At the bottom layer, $\hb_{0}$ is the input $\xb$. The
top hidden layer $\hb_{T}$ is connected to the outcome.

Training very deep feedforward networks remains difficult for several
reasons. First, the number of parameters grows with the depth of the
network, which leads to overfitting. Second, the stack of multiple
non-linear functions makes it difficult for the information and the
gradients to pass through.

In Highway Networks, there are two modifications that resolve these
problems: (i) Parameters are shared between layers leading to a compact
model, and (ii) The activation function is modified by adding sigmoid
gates that let information from lower layers pass linearly through.
Fig.~\ref{fig:highway_GRU}(a) illustrates a Highway Network layer.
The first modification requires that all the hidden layers to have
the same hidden units $k$. The bottom layer is identical to that
of standard feedforward networks. The second modification defines
a candidate hidden state $\tilde{\hb}_{t}\in\mathbb{R}^{k}$ as the
usual non-linear transform:

\[
\tilde{\hb}_{t}=g\left(W\hb_{t-1}+\bb\right)
\]
where $W$ and $\bb$ are parameter matrix and bias vector that shared
among all hidden layers. Finally the hidden state is gated by two
gates $\alphab_{1},\alphab_{2}\in\left[0,1\right]^{k}$ as follows:
\begin{eqnarray}
\hb_{t} & = & \alphab_{1}*\tilde{\hb}_{t}+\alphab_{2}*\hb_{t-1}\label{eq:gated-update}
\end{eqnarray}
for $t\geq2$. The two gates $\alphab_{1}$ and $\alphab_{2}$ are
sigmoid functions and can be independent, where $\alphab_{1}=\sigma\left(U_{1}\hb_{t-1}+\cb_{1}\right)$
and $\alphab_{2}=\sigma\left(U_{2}\hb_{t-1}+\cb_{2}\right)$ or summed
to unit element-wise, e.g., $\mathbf{1}=\alphab_{1}+\alphab_{2}$.
The latter option was used in the paper of Highway Networks \cite{srivastava2015training}.

The part $\alphab_{2}*\hb_{t-1}$, which is called \emph{carry behavior},
makes the information from layers below pass easily through the network.
This behavior also allows the back-propagation to compute the gradient
more directly to the input. The net effect is that the networks can
be very deep (up to thousand of layers).

\subsection{Gated Recurrent Units \label{subsec:Gated-Recurrent-Units}}

\subsubsection*{Recurrent neural networks}

A recurrent neural network (RNN) is an extension of feedforward networks
to map a variable-length input sequence $\xb_{1},...,\xb_{T}$ to
a output sequence $y_{1},...,y_{T}$. An RNN allows self-loop connections
and shared parameters across all steps of the sequence. For vanilla
RNNs, the activation (which is also called hidden state) $\hb_{t}$
is a function of the current input and the previous hidden state $\hb_{t-1}$:

\begin{equation}
\hb_{t}=g\left(W\xb_{t}+U\hb_{t-1}+\bb\right)\label{eq:RNN-unit}
\end{equation}
where $W,U$ and $\bb$ are parameters shared among all steps. However,
vanilla RNNs suffer from vanishing gradient for large $T$, thus preventing
the use for long sequences \cite{hochreiter2001gradient}.

\subsubsection*{Gated Recurrent Units}

A Gated Recurrent Unit (GRU) \cite{cho2014learning,cho2014properties}
is an extension of vanilla RNNs (see Fig.~\ref{fig:highway_GRU}(b)
for an illustration) that does not suffer from the vanishing gradient
problem. At each step $t$, we fist compute a candidate hidden state
$\tilde{\hb}_{t}$ as follows:

\begin{eqnarray*}
\rb_{t} & = & \sigma\left(W_{r}\xb_{t}+U_{r}\hb_{t-1}+\bb_{r}\right)\\
\tilde{\hb}_{t} & = & \mbox{tanh}\left(W_{h}\xb_{t}+U_{h}\left(\rb_{t}*\hb_{t-1}\right)+\bb_{h}\right)
\end{eqnarray*}
where $\rb_{t}$ is a reset gate that controls the information flow
of the previous state to the candidate hidden state. When $\rb_{t}$
is close to $\mathbf{0}$, the previous hidden state is ignored and
the candidate hidden state is reset with the current input.

GRUs then update the hidden state $\hb_{t}$ using the same rule as
in Eq.~(\ref{eq:gated-update}). The difference is in the gate function:
$\alphab_{1}=\sigma\left(W_{\alphab}\xb_{t}+U_{\alphab}\hb_{t-1}+\bb_{\alphab}\right)$,
where current input $\xb_{t}$ is used. The linear relationship between
the two gates is assumed: $\alphab_{2}=1-\alphab_{1}$. This relationship
enables the hidden state from previous step to be copied partly to
the current step. Hence $\hb_{t}$ is a linear interpolation of the
candidate hidden state and all previous hidden states. This prevents
the gradients from vanishing and captures longer dependencies in the
input sequence.

\subsubsection*{Remark}

Both Highway Networks and GRUs can be considered as simplified versions
of Long Short-Term Memory \cite{hochreiter1997long}. With the linear
interpolation between consecutive states, the GRUs have less parameters.
Empirical experiments revealed that GRUs are comparable to LSTM and
more efficient in training \cite{chung2014empirical}.

\subsection{$p$-norm Gates \label{subsec:Pointwise-unit-norms}}

As described in the two sections above, the gates of the non-linear
and linear parts in both Highway Networks (the version empirically
validated in \cite{srivastava2015training}) and GRUs use the same
linear constraint:
\[
\alphab_{1}+\alphab_{2}=\mathbf{1},\quad\mbox{s.t.}\quad\alphab_{1},\alphab_{2}\in\left(0,1\right)^{k}
\]
where $\alphab_{1}$ plays the updating role and $\alphab_{2}$ plays
the forgetting role in the computational sequence. Since the relationship
is linear, when $\alphab_{1}$ gets closer to $\mathbf{1}$, $\alphab_{2}$
will get closer to $\mathbf{0}$ at the same rate. During learning,
the gates might become more specialized and discriminative, this same-rate
convergence may block the information from the lower layer passing
through at a high rate. The learning speed may suffer as a result.

We propose to relax this scheme by using the following the following
$p$-norm scheme:
\begin{equation}
\left(\alphab_{1}^{p}+\alphab_{2}^{p}\right)^{\frac{1}{p}}=\mathbf{1},\quad\mbox{equivalently:}\quad\alphab_{2}=\left(\mathbf{1}-\alphab_{1}^{p}\right)^{\frac{1}{p}}\label{eq:p-norm}
\end{equation}
for $p>0$, where the norm is applied element-wise. 

The dynamics of the relationship of the two gates as a function of
$p$ is interesting. For $p>1$ we have $\alphab_{1}+\alphab_{2}>\mathbf{1}$.
This increases the amount of information passing for the linear part.
To be more concrete, let $\alphab_{1}=\mathbf{0.9}$. For the linear
gates relationship with $p=1$, there is a portion of $\alphab_{2}=\mathbf{0.1}$
of old information passing through each step. But for $p=2$, the
passing portion is $\alphab_{2}=\mathbf{0.4359}$, and for $p=5$,
it is $\alphab_{2}=\mathbf{0.865}$. When $p\rightarrow\infty$, $\alphab_{2}\rightarrow\mathbf{1}$,
regardless of $\alphab_{1}$ as long as $\alphab_{1}<\mathbf{1}$.
This is achievable since $\alphab_{1}$ is often modelled as a logistic
function. When $p\rightarrow\infty$, $\alphab_{2}\rightarrow\mathbf{1}$,
the activation of the final hidden layer loads all the information
of the past without forgetting. Note that the ImageNet winner 2015,
the Residual Network \cite{he2015deep}, is a special case with $\alphab_{1},\alphab_{2}\rightarrow\mathbf{1}$. 

On the other hand, $p<1$ implies $\alphab_{1}+\alphab_{2}<\mathbf{1}$,
the linearity gates are closed at a faster rate, which may prevent
information and gradient flow passing easily through layers.

\section{Experiments}

In this section, we empirically study the behavior of the $p$-norm
gates in feedforward networks (in particular, Highway Networks presented
in Section.~\ref{subsec:Highway-Networks}) and recurrent networks
(Gated Recurrent Unit in Section.~\ref{subsec:Gated-Recurrent-Units}).

\subsection{Vector Data with Highway Network}

We used vector-data classification tasks to evaluate the Highway Networks
under $p$-norm gates. We used 10 hidden layers of 50 dimensions each.
The models were trained using standard stochastic gradient descent
(SGD) for 100 epochs with mini-batch of 20.

\subsubsection*{Datasets}

We used two large UCI datasets: MiniBooNE particle identification
(MiniBoo) \footnote{https://archive.ics.uci.edu/ml/datasets/MiniBooNE+particle+identification}
and Sensorless Drive Diagnosis (Sensorless) \footnote{https://archive.ics.uci.edu/ml/datasets/Dataset+for+Sensorless+Drive+Diagnosis}.
The first is a binary classification task where data were taken from
the MiniBooNE experiment and used to classify electron neutrinos (signal)
from muon neutrinos (background). The second dataset was extracted
from motor current with 11 different classes. Table~1 reports the
data statistics.

\begin{table}[h]
\caption{Datasets for testing Highway Networks.}

\centering{}%
\begin{tabular}{|c|c|c|c|c|}
\hline 
Dataset & Dimens. & Classes & Train. set & Valid. set\tabularnewline
\hline 
\hline 
MiniBoo & 50 & 2 & 48,700 & 12,200\tabularnewline
\hline 
Sensorless & 48 & 11 & 39,000 & 9,800\tabularnewline
\hline 
\end{tabular}
\end{table}

\subsubsection*{Training curves}

\begin{figure*}[!t]
\begin{centering}
\begin{tabular}{cc}
\includegraphics[width=0.45\textwidth]{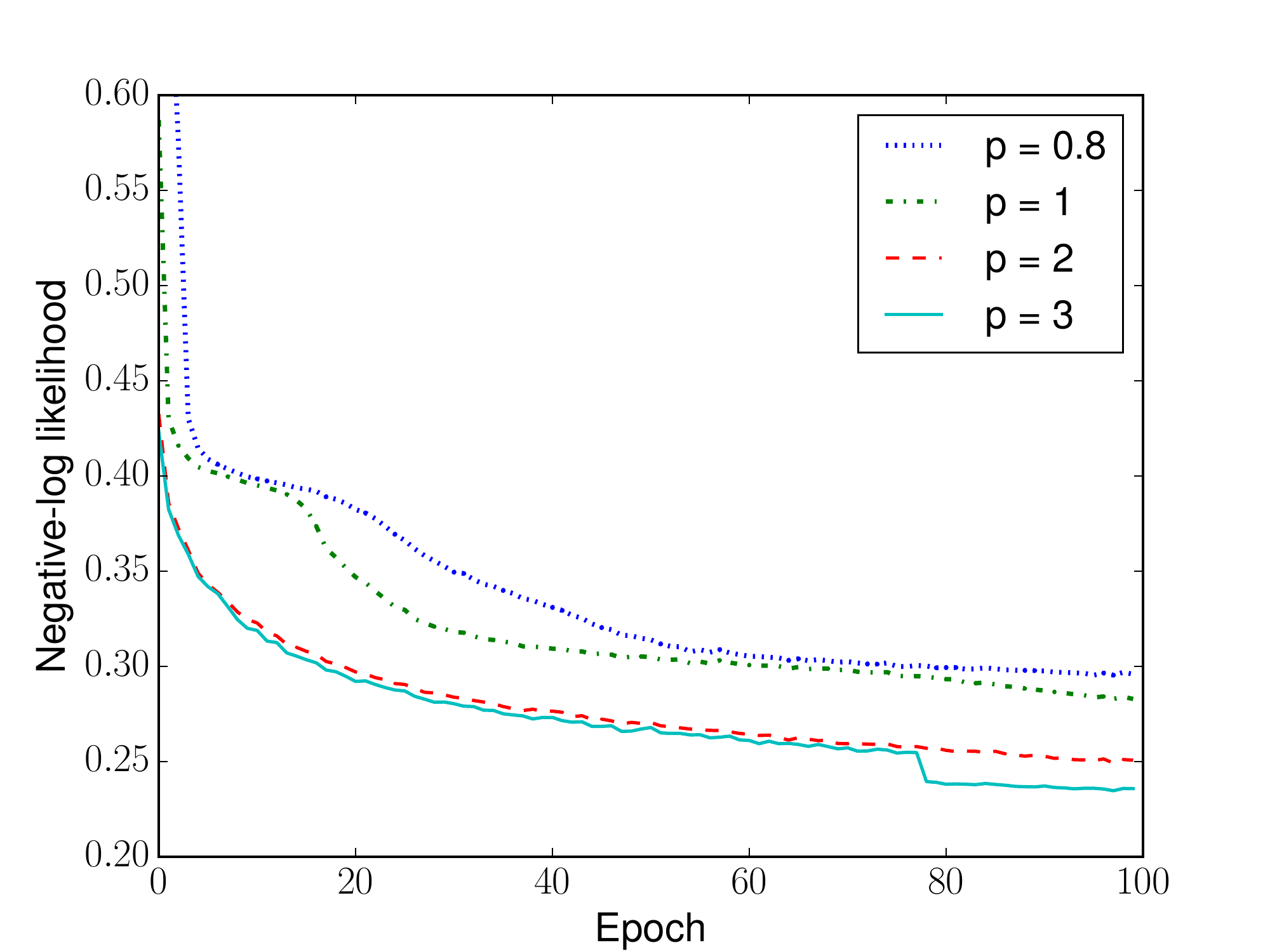} & \includegraphics[width=0.45\textwidth]{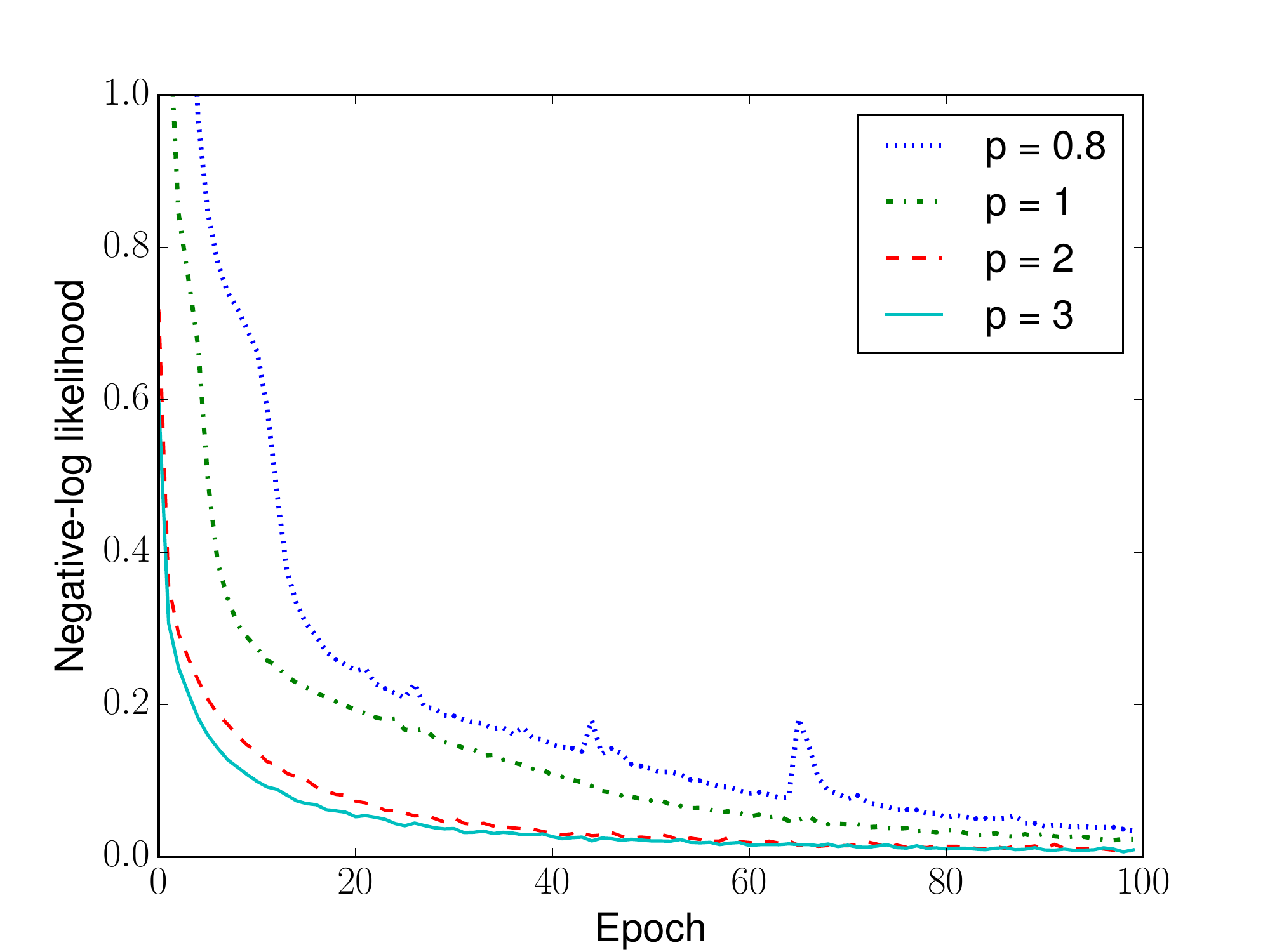}\tabularnewline
(a) & (b)\tabularnewline
\end{tabular}
\par\end{centering}
\caption{Learning curves on training sets. (a) MiniBoo dataset. (b) Sensorless
dataset.\label{fig:Learning-curves-highway}}
\end{figure*}

Fig.~\ref{fig:Learning-curves-highway} shows the training curves
on training sets. The loss function is measured by negative-log likelihood.
The training costs with $p=2$ and $p=3$ decrease and converge much
faster than ones with $p=0.8$ and $p=1$. In the MiniBoo dataset,
training with $p=2$ and $p=3$ only needs 20 epochs to reach 0.3
nats, while $p=1$ needs nearly 100 epochs and $p=0.8$ does not reach
that value. The pattern is similar in the Sensorless dataset, the
training loss for $p=1$ is 0.023 after 100 epochs, while for $p=2$
and $p=3$, the losses reach that value after 53 and 44 epochs, respectively.
The training for $p=0.5$ was largely unsuccessful so we do not report
here.

\subsubsection*{Prediction}

The prediction results on the validation sets are reported in Table~\ref{tab:Results-highway}.
To evaluate the learning speed, we report the number of training epochs
to reach a certain benchmark with different values of $p$. We also
report the results after 100 epochs. For the MiniBoo dataset (Table~\ref{tab:Results-highway}(a)),
$p=0.8$ does not reach the benchmark of 89\% of F1-score, $p=1$
needs 94 epochs while both $p=2$ and $p=3$ need 33 epochs, nearly
3 times faster. For the Sensorless dataset (Table~\ref{tab:Results-highway}(b)),
$p=3$ has the best result and needs only 35 epochs to achieve 99\%
of macro F1-score while $p=1$ and $p=2$ need 77 and 41 epochs, respectively.

\begin{table*}[!t]
\caption{\label{tab:Results-highway}Results on validation sets. The second
column is the number of epochs to reach a benchmark measured by F1-score
(\%) for MiniBoo and macro F1-score (\%) for Sensorless. The third
column is the results after running 100 epochs.}

\centering{}%
\begin{tabular}{cc}
(a) MiniBoo dataset & (b) Sensorless dataset\tabularnewline
\begin{tabular}{|c|c|c|}
\hline 
$p$ & epochs to 89\% F1-score & F1-score (\%)\tabularnewline
\hline 
\hline 
0.8 & N/A & 88.5\tabularnewline
1 & 94 & 89.1\tabularnewline
2 & \textbf{33} & 90.2\tabularnewline
3 & \textbf{33} & \textbf{90.4}\tabularnewline
\hline 
\end{tabular} & %
\begin{tabular}{|c|c|c|}
\hline 
$p$ & epochs to 99\% F1-score & macro F1-score (\%)\tabularnewline
\hline 
\hline 
0.8 & 92 & 99.1\tabularnewline
1 & 77 & 99.4\tabularnewline
2 & 41 & 99.7\tabularnewline
3 & \textbf{35} & \textbf{99.7}\tabularnewline
\hline 
\end{tabular}\tabularnewline
\end{tabular}
\end{table*}

\subsubsection*{Visualization}

\begin{figure*}[!t]
\centering{}%
\begin{tabular}{c}
\includegraphics[bb=50bp 0bp 1150bp 360bp,clip,width=0.95\textwidth,height=0.28\textwidth]{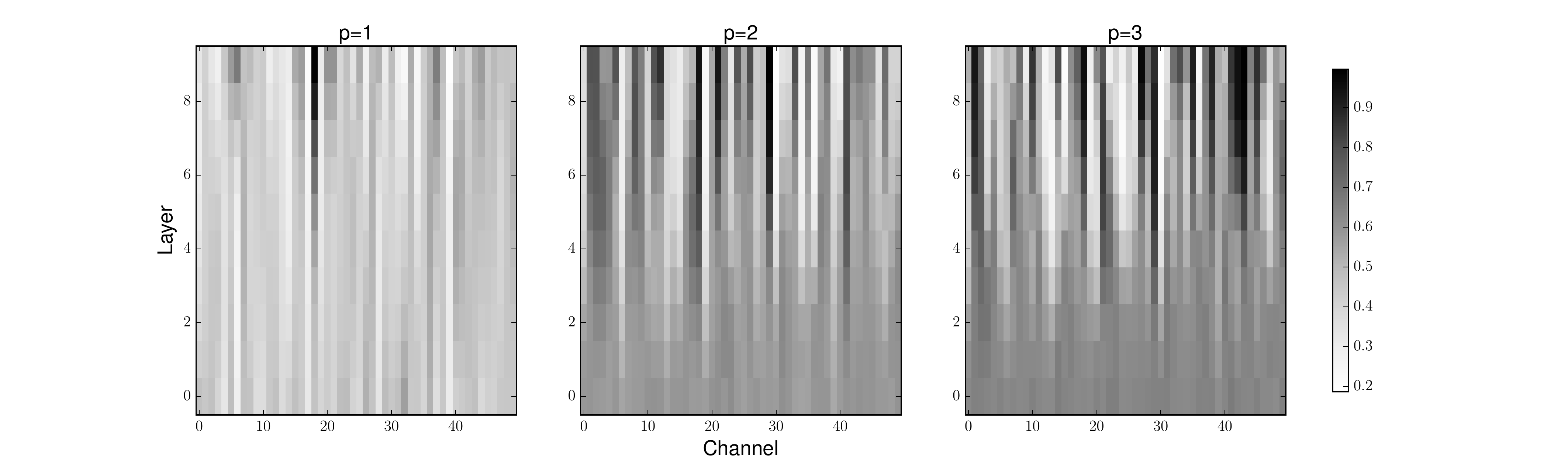}\tabularnewline
(a) The gate $\alphab_{1}$\tabularnewline
\includegraphics[bb=50bp 0bp 1150bp 360bp,clip,width=0.95\textwidth,height=0.28\textwidth]{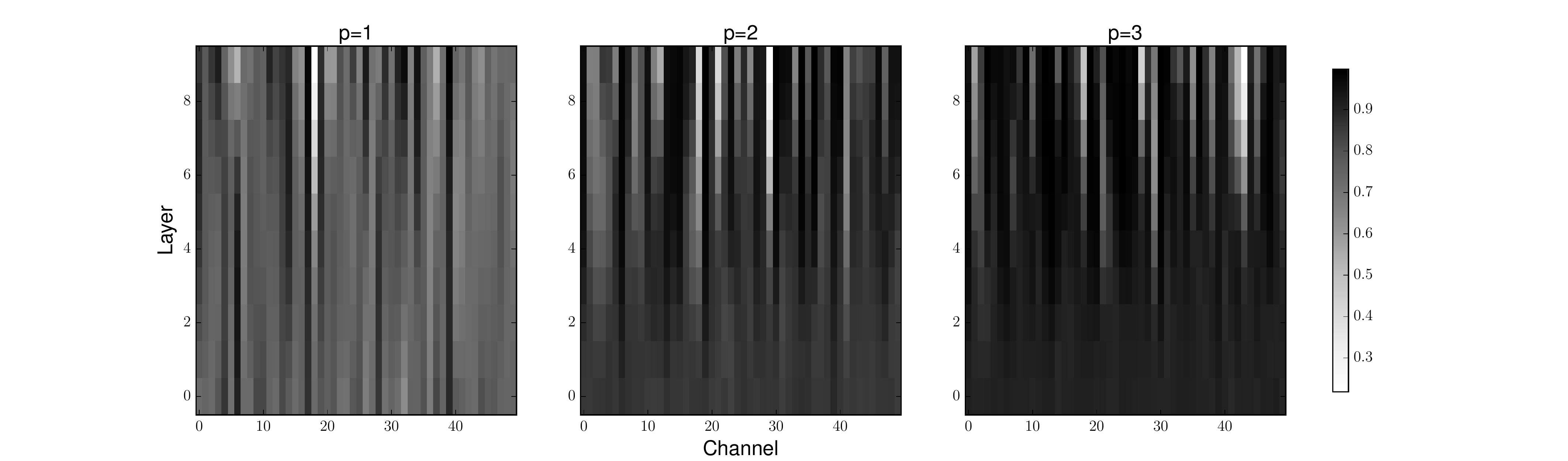}\tabularnewline
(b) The gate $\alphab_{2}$\tabularnewline
\end{tabular}\caption{The dynamics of 50 channels of the two gates through 10 layers with
different $p$. \label{fig:Gate_opening}}
\end{figure*}

Fig.~\ref{fig:Gate_opening} illustrates how 50 channels of the two
gates open through 10 layers with different value of $p$ for a randomly
chosen data instance in the test set of MiniBoo. Recall from Sec.~\ref{subsec:Pointwise-unit-norms}
that $\alphab_{1}$ and $\alphab_{2}$ control the amount of information
in the non-linearity part and the linearity part, respectively and
$\alphab_{1}^{p}+\alphab_{2}^{p}=\mathbf{1}$. It is clear that with
the larger value of $p$, the more the two gates are open. Interestingly,
the values of most channels in the gate $\alphab_{2}$ are larger
than ones in the gate $\alphab_{1}$ for all values of $p$. The model
seems to prefer the linearity part. More interestingly, there is a
gradual change in gates over the layers, although gates between layers
are not directly linked. At the lower layers, gates are more uniform,
but get more informative near the top (which is closer to the outcome).

\subsection{Sequential Data with GRUs}

To evaluate the $p$-norm with GRUs, we compare the results with different
values of $p$ in the task of language modeling at character-level
\cite{graves2013generating,mikolov2012subword,kim2015character}.
This modeling level has attracted a great interest recently due to
their generalizability over language usage, a very compact state space
(which equals the alphabet size, hence the learning speed), and the
ability to capture sub-word structures. More specifically, given a
sequence of characters $c_{1},...,c_{t-1}$, the GRU models the probability
of the next character $c_{t}$: $P\left(c_{t}\mid c_{t-1},...,c_{1}\right)$.
The quality of model is measured by \emph{bits-per-character} in the
test set which is $-\mbox{log}_{2}P\left(c_{t}\mid c_{t-1},...,c_{1}\right)$.
The model is trained by maximizing the log-likelihood: $\sum_{t=1}^{T}\log P\left(c_{t}\mid c_{t-1},...,c_{1}\right)$.

\subsubsection*{Dataset}

We used the UCI Reuters dataset which contains articles of 50 authors
in online Writeprint \footnote{https://archive.ics.uci.edu/ml/datasets/Reuter\_50\_50}.
We randomly chose $10,000$ sentences for training and $4,000$ sentences
for validation. For sentences with length longer than 100 characters,
we only used the first 100 characters. The model is trained with 400
hidden units, 50 epochs and 32 sentences each mini-batch.

\subsubsection*{Results}

\begin{figure*}[!t]
\begin{centering}
\begin{tabular}{cc}
\includegraphics[width=0.45\textwidth]{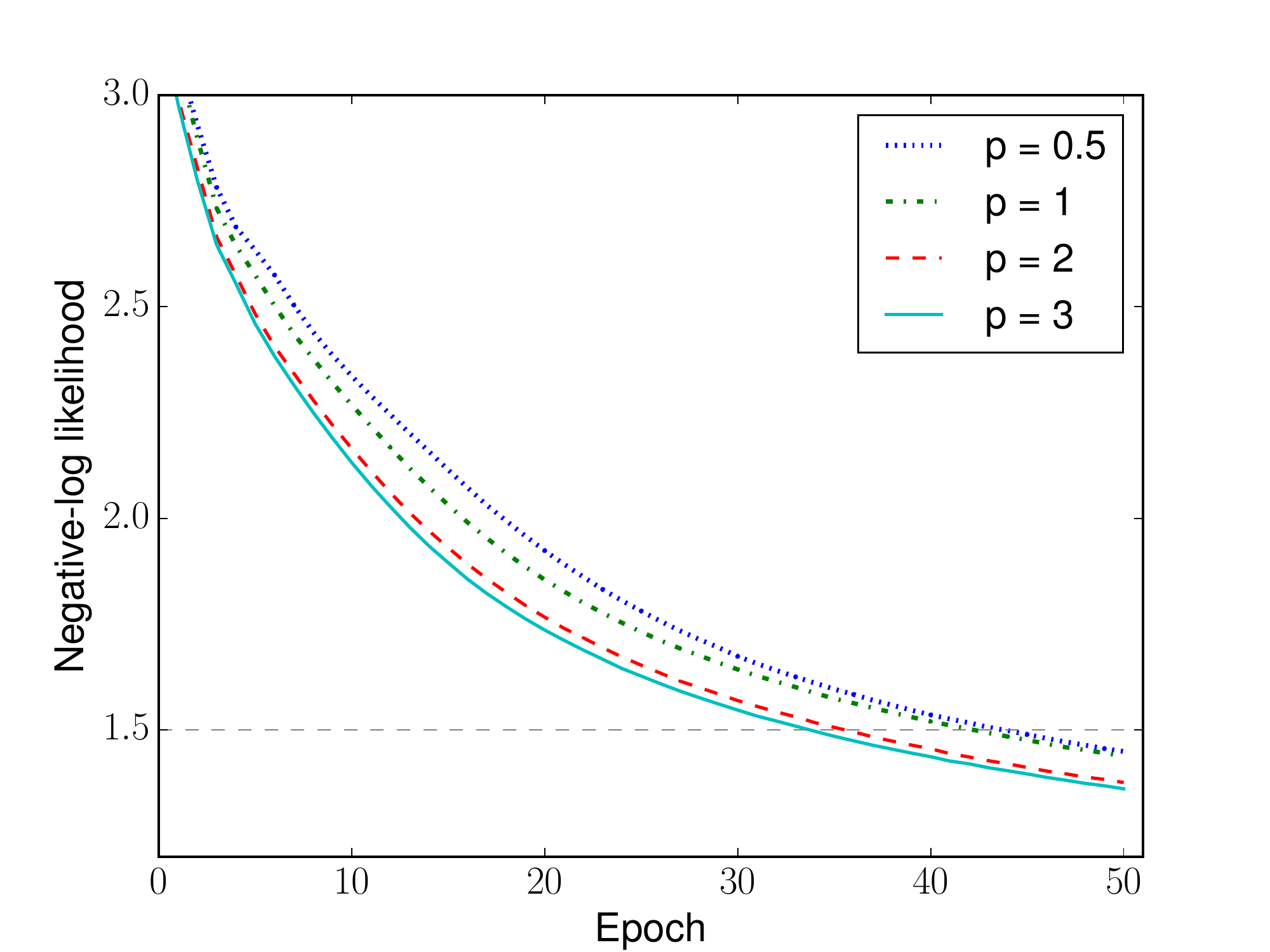} & \includegraphics[width=0.45\textwidth]{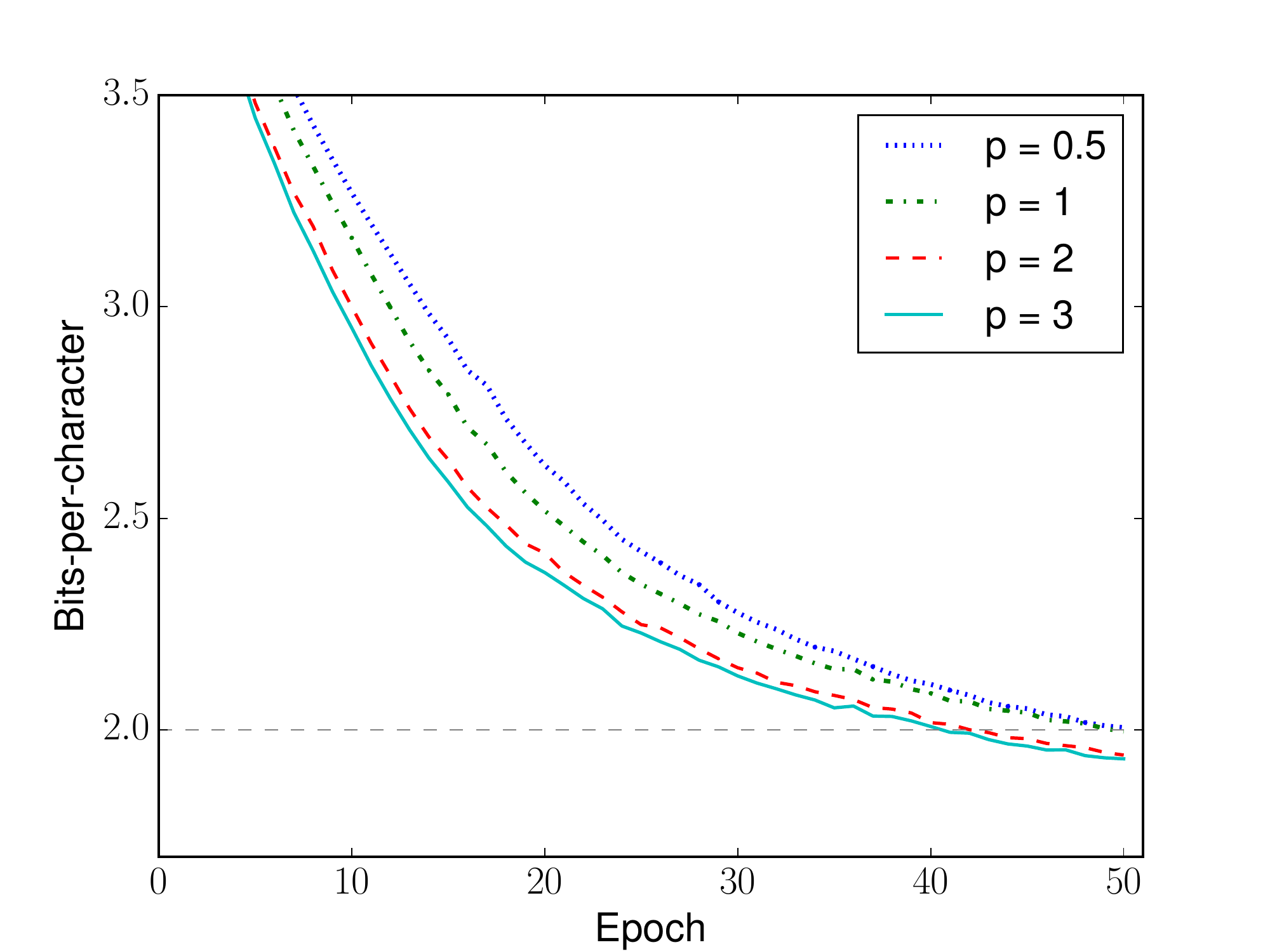}\tabularnewline
(a) & (b)\tabularnewline
\end{tabular}
\par\end{centering}
\caption{(a). Learning curve on training set. (b). Results on validation set
measured by Bits-per-character \label{fig:GRUs_results}}
\end{figure*}

Fig.~\ref{fig:GRUs_results} reports (a) training curves and (b)
results on validation set through epochs. It is clear from the two
figures that the model with $p=3$ performs best among the choices
$p\in\left(0.5,1,2,3\right)$, both in learning speed and model quality.
To give a more concrete example, as indicated by the horizontal lines
in Figs.~\ref{fig:GRUs_results}(a,b), learning with $p=3$ reaches
1.5 nats after 34 epoch, while learning $p=1$ reaches that training
loss after 43. For model quality on test data, model with $p=3$ achieves
2.0 bits-per-character after 41 epochs, faster than model with $p=1$
after 50 epochs.

\subsection{Evaluating the Effectiveness of $p$\label{subsec:efficiency_of_p}}

We have demonstrated in both Highway Nets and GRUs that training is
faster with $p>1$. However, we question whether the larger value
of $p$ always implies the better results and faster training? For
example, as $p\rightarrow\infty$, we have $\alphab_{2}\rightarrow1$
and the activation at the final hidden layer contains a copy of the
first layer and all other candidate states: $\hb_{T}=\hb_{1}+\sum_{t=2}^{T}\alphab_{t}\tilde{\hb}_{t}$.
This makes the magnitude of hidden states not properly controllable
in very deeper networks. To evaluate the effectiveness of $p$, we
conducted experiments on the MiniBoo dataset with $p=0.8,1,2,...,8$
and networks with depths of 10, 20, 30. We found that the model works
very well for all values of $p$ with 10 hidden layers. When the number
of layers increase (let say 20 or 30), the model only works well with
$p=2$ and $p=3$. This suggests that a proper control of the hidden
state norms may be needed for very deep networks and widely open gates.

\begin{figure}
\begin{centering}
\includegraphics[width=0.95\columnwidth]{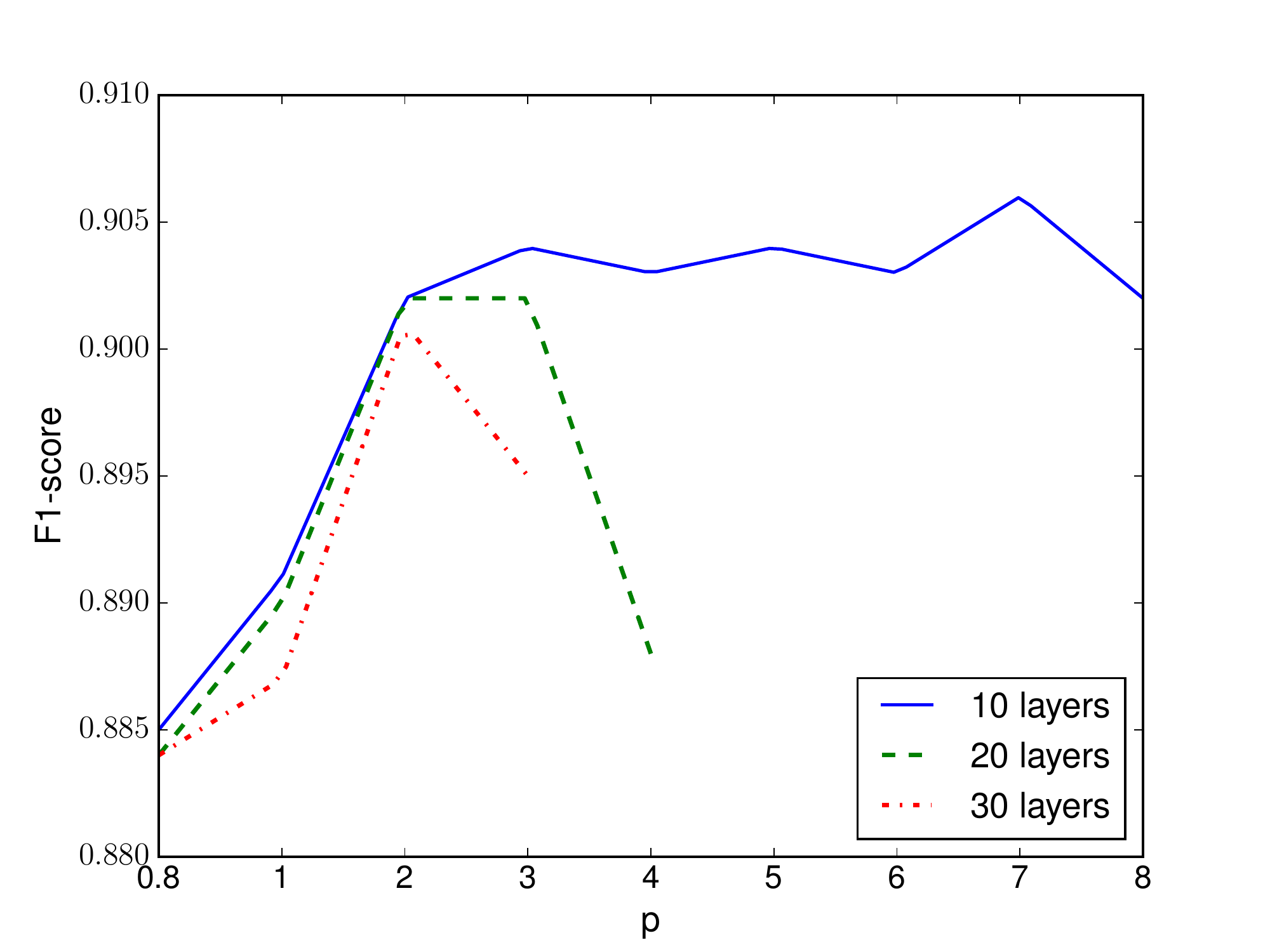}
\par\end{centering}
\caption{Results on Miniboo dataset with different $p$ and different number
of layers}

\end{figure}

\section{Discussion and Conclusion}

\subsection{Discussion}

Gating is a method for controlling the linearity of the functions
approximated by deep networks. Another method is to use piece-wise
linear units, such as those in ReLU family \cite{glorot2011deep,he2015delving,goodfellow2013maxout}.
Still, partial or piece-wise linearity has  desirable nonlinearity
for complex functions. At the same time it helps to prevent activation
units from being saturated and the gradients from being vanishing,
making gradient-based learning possible for very deep networks \cite{srivastava2015training,hochreiter1997long}.
The main idea of $p$-norm gates is to allow a greater flow of data
signals and gradients through many computational steps. This leads
to faster learning, as we have demonstrated through experiments.

It remains less clear about the dynamics of the relationship between
the linearity gate $\alphab_{2}$ and nonlinearity gate $\alphab_{1}$.
We hypothesize that, at least during the earlier stage of the learning,
larger gates help to improve the credit assignment by allowing easier
gradient communication from the outcome error to each unit. Since
the gates are learnable, the amount of linearity in the function approximator
is controlled automatically. 

\subsection{Conclusion}

In this paper, we have introduced $p$-norm gates, a flexible gating
scheme that relaxes the relationship between nonlinearity and linearity
gates in state-of-the-art deep networks such as Highway Networks,
Residual Networks and GRUs. The $p$-norm gates make the gates generally
wider for larger $p$, and thus increase the amount of information
and gradient flow passing through the networks. We have demonstrated
the $p$-norm gates on two major settings: vector classification tasks
with Highway Networks and sequence modelling with GRUs. The extensive
experiments consistently demonstrated that faster learning is caused
by $p>1$.

There may be other ways to control linearity through the relationship
between the linearity gate $\alphab_{2}$ and nonlinearity gate $\alphab_{1}$.
A possible scheme could be a monotonic relationship between the two
gates so as $\alphab_{1}\rightarrow0$ then $\alphab_{2}\rightarrow1$
and $\alphab_{1}\rightarrow1$ then $\alphab_{2}\rightarrow0$. It
also remains open to validate this idea on LSTM memory cells, which
may lead to a more compact model with less than one gate parameter
set. The other open direction is to modify the internal working of
gates to make them more informative \cite{pham2016deepcare}, and
to assist in regularizing the hidden states, following the findings
in Sec.~\ref{subsec:efficiency_of_p} and also in a recent work \cite{krueger2015regularizing}.

\bibliographystyle{IEEEtran}
%\bibliography{../../bibs/ME,../../bibs/truyen,../../bibs/trang}


\end{document}